\documentclass[preprint,3p,12pt]{elsarticle}
\usepackage{verbatim}
\usepackage{amsmath,bm}

\usepackage[utf8]{inputenc} 
\usepackage[T1]{fontenc}    
\usepackage{hyperref}       
\usepackage{url}            
\usepackage{booktabs}       
\usepackage{amsfonts}       
\usepackage{nicefrac}       
\usepackage{microtype}      
\usepackage{verbatim}
\usepackage{epsfig}
\usepackage{epsf}
\usepackage{epstopdf}
\usepackage{graphicx}
\usepackage{subfigure}
\usepackage{amsmath}
\usepackage{bigints}

\title{{\bf {\large PH-VAE: A Polynomial Hierarchical Variational Autoencoder Towards Disentangled Representation Learning}}}

\author{%
  Xi Chen\textsuperscript{1,2}, Shaofan Li \textsuperscript{1,$\dagger$} \\
  \textsuperscript{1} Department of Civil and Environmental Engineering, \\
    University of California, Berkeley, CA, 94720, USA\\
  \textsuperscript{2} Key Laboratory of Soft Machines and Smart Devices of Zhejiang Province, \\
Department of Engineering Mechanics, Zhejiang University, Hangzhou, 310027, China\\
\textsuperscript{$\dagger$} Correspondence: \texttt{shaofan@berkeley.edu} \\
}

\begin{document}

\begin{abstract}
The variational autoencoder (VAE) is a simple and efficient
generative artificial intelligence
method for modeling complex probability distributions of various
types of data, including images and texts.
However, it suffers some main shortcomings, such as
lack of interpretability in the latent variables,
difficulties in tuning hyperparameters while training,
producing blurry, unrealistic downstream outputs or loss
of information due to how it calculates loss functions
and recovers data distributions, overfitting,
and origin gravity effect for small data sets, among other issues.
These and other limitations have caused
unsatisfactory generation effects for the data with complex distributions.
In this work, we proposed and developed a polynomial hierarchical
variational autoencoder (PH-VAE), in which we used a polynomial hierarchical
date format to generate or to reconstruct the data distributions.
In doing so, we also proposed
a novel Polynomial Divergence in the loss function to replace or generalize
the Kullback-Leibler (KL) divergence,
which results in systematic and drastic improvements in both
accuracy and reproducibility of the re-constructed distribution function
as well as the quality of re-constructed data images
while keeping the dataset size the same but capturing fine resolution of
the data. Moreover, we showed that the proposed PH-VAE has some
disentangled representation learning ability to distinguish and decompose
information or data at different scales.
\end{abstract}

\maketitle


\vskip 0.5in

\section{Introduction}
Variational autoencoders (VAEs) are simple and powerful generative
 artificial intelligence models that use
 deep learning to generate new data content, detect existing data anomalies and hence remove noise
 or unwanted data \citep{Cemgil2020, wei2020, Zemouri2022, Singh2021}.
 These technical capabilities make VAEs well suited for numerous powerful applications,
such as image and text creations, video creation, Synthetic data creation, language processing,
and existing data anomaly detection or anomaly detection in  time series or sequential data processing.
Moreover, VAE is able to predict how much noise to add to each embedding
during training stage before it is passed to the decoder. It makes VAE
able to explicitly learn how to model the 'gaps' between the training datapoints,
therefore giving better results for generating novel samples or data points.
In the recent decade, it has been extensively applied to many scientific
and engineering problems, e.g. \citep{Liu2023}.

However, the original VAE suffers some serious limitations, especially poor reconstruction, as well as
generation effects for data with complex distribution, gravitating towards
to center, insufficient disentangling effect, difficulties in training and
tuning hyperparameters, etc. among others.
To resolve these issues,
various improved version of VAEs have been proposed, including $\beta$-VAE \citep{Higgins2017, Burgess2018},
hierarchical VAEs, e.g. \citep{Vahdat2020}, DIP-VAE \citep{Kumar2017}, Info-VAE \citep{Zhao2019},
the autoencoding-VAE \citep{An2015}, Factor VAE \citep{Kim2018,Kim2019},
TC-VAE \citep{Noseworthy2020} and
the dynamical variational autoencoders (DVAEs) \citep{Girin2020}, among
many others. Nevertheless, the lack of
meaningful mathematically precise definition of disentanglement remains
difficult to find, and hence, the lack of reproducibility of VAEs remains
for data sets with complex distributions e.g. \citep{yacoby2020}.

In original VAE and some of its variants, the learned latent space
 is typically disorganized and entangled, and the construction of
 the latent space tends to homogenize or isotropize the anisotropic
 data features.
 One of the approaches to recover the data diversity and its richness in texture is
 to utilize the disentangled representation learning (DRL),
  which is a learning model capable of identifying and disentangling the underlying factors
  hidden in the observable data in representation form \cite{Wang2024,Pastrana2022}.
  However, to do this in an unsupervised manner is still a great challenge
  in generative artificial intelligence models.
  Even some of the above-mentioned VAE models have more or less addressed this issue.
  The fundamental problem persisted. This is because, in a unsupervised learning,
  we do not have an effective means
  to identify and separate the underlying factors in data.

In this work, we approach this problem from a complete different angle
by first disentangle the input features into a polynomial hierarchical form,
without increasing the data size, and
we then introduce a novel polynomial divergence into the loss function
of the proposed polynomial hierarchical variational autoencorder (PH-VAE).
By doing so, we showed the proposed PH-VAE can provide a
robust unsupervised disentangled learning (See: \cite{Wang2024,Mao2024})
with enhanced data reproducibility and data generation ability.

\bigskip
\begin{figure}[h]
		\begin{center}
			\includegraphics[width=5.0in]{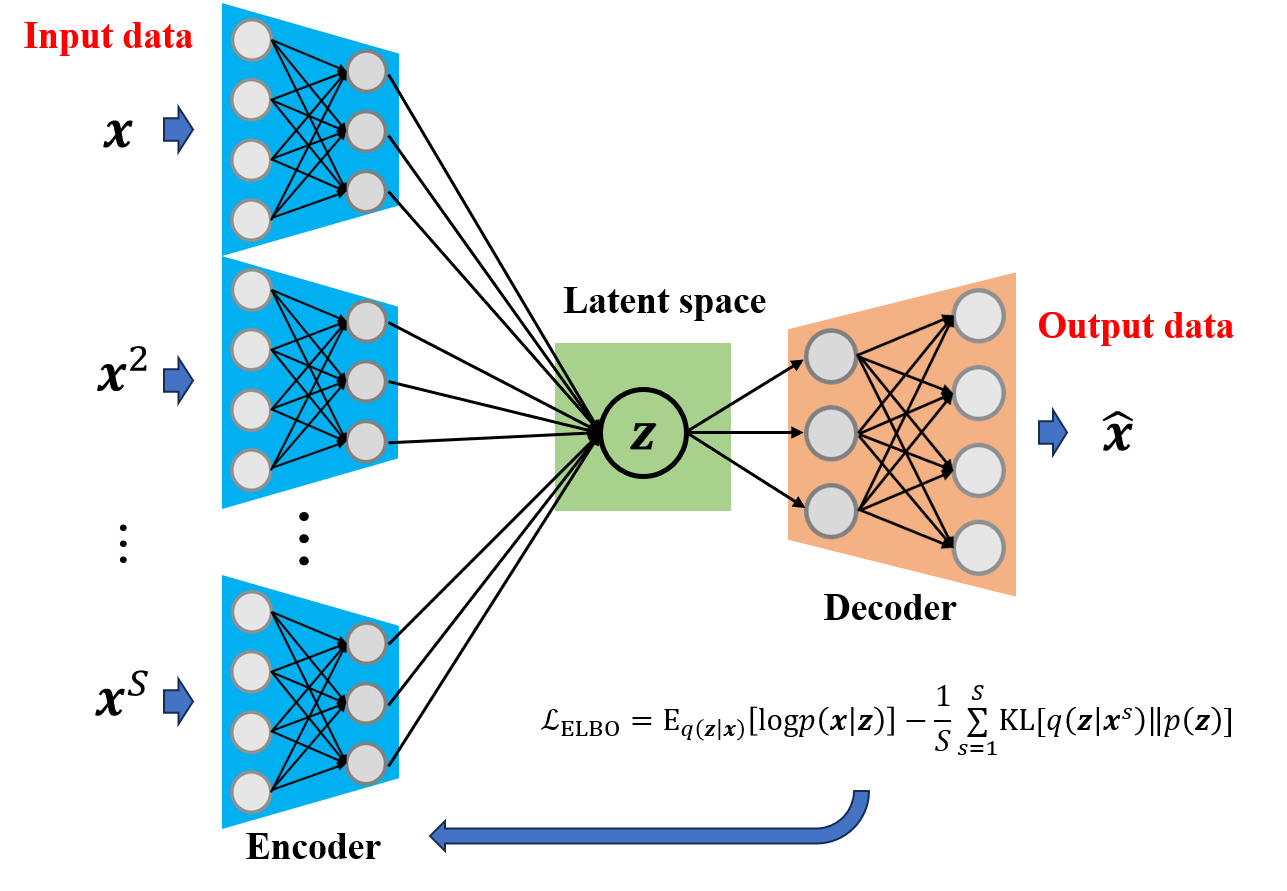}
		\end{center}
	\caption{Architecture of the polynomial-enriched hierarchical variational autoencoder.}
	\label{fig:concrete-fig1a}
\end{figure}

\section{Theoretical Formulation and Computer Implementation}
 In this section, a polynomial-enriched hierarchical variational autoencoder (PH-VAE) framework is proposed, which simultaneously processes multiple encoder layers yet integrates their outputs into a unified decoder. The design of PH-VAE is dedicated to addressing two key limitations: enhancing reconstitution precision and enriching the latent space with more information to mitigate posterior collapse. We then outline the theoretical foundation and implementation details of PH-VAE.

\subsection{Data preprocessing and transformation}

 The data samples, originating from either experimental measurements or real-world applications, form the foundation for the analysis or model development process. Assuming that raw data is structured as follows,

  \begin{equation}
  \{\boldsymbol{y}_n\} \quad (n = 1, \dots, N),
 \end{equation}

\noindent
 where $n$ is the number of features; $\boldsymbol{y}_n$ denotes the sample vector regarding each feature $n$. The dataset adopted for PH-VAE involves standardizing raw data to a consistent scale. Herein, all data are mapped to the range $[0, 1]$ as follows,

  \begin{equation}
  \boldsymbol{x}_n = \frac{\boldsymbol{y}_n - \min(\boldsymbol{y}_n)}{\max(\boldsymbol{y}_n) - \min(\boldsymbol{y}_n)},
 \end{equation}
 where $\min(\cdot)$ represents the minimum value of $\cdot$ and $\max(\cdot)$ represents the maximum value. While for image data, additional steps are required, comprising resizing the images to a prescribed dimension and applying enhancement techniques to increase the model's generalization.

 In order to extract more information without requiring additional sampling,
 the dataset \( \boldsymbol{x}_n \) undergoes a series of transformations to generate polynomial features, such as the original data \( \boldsymbol{x}_n^1 \), its square \( \boldsymbol{x}_n^2 \), its cube \( \boldsymbol{x}_n^3 \), and so on. These higher-order features enable the model to capture more complex relationships within the data, improving its capacity to learn non-linear patterns. The transformed data is then transposed to ensure each row represents a feature vector corresponding to a single sample, following the standard format used in PH-VAE models. Therefore, the dataset is provided as follows,

 \begin{equation}
  \{\boldsymbol{x}_n^s\} \quad (n = 1, \dots, N;s = 1, \dots, S),
 \end{equation}

\noindent
 where $s$ is an index to denote the $n$-th data set of the total $N$ data set,
 and each data set has $m$-number of feature.
 The index $s$ indicates $s$-th power operation on the original data.

 For example, for a given input data feature set, ${\boldsymbol{x}}_n^T = (x_1, x_2, \cdots, x_m)_n$,
 we can reformat it as
 \begin{equation}
 {\boldsymbol{x}}_n ~\to ~\left[
 \begin{array}{c}
 {\boldsymbol{x}}^1_n\\
 {\boldsymbol{x}}_n^2 \\
 \vdots\\
 {\boldsymbol{x}}_n^s\\
 \vdots\\
 {\boldsymbol{x}}_n^S\\
 \end{array}
 \right],
 \end{equation}
 where ${\boldsymbol{x}}^s$ is defined as
 \begin{equation}
 {\boldsymbol{x}}^s_n :=
 \left(
 \begin{array}{c}
 x_{n1}^s\\
 x_{n2}^s \\
 \vdots\\
 x_{nj}^s\\
 \vdots\\
 x_{nm}^s\\
 \end{array}
 \right),~~~~s=1,2, \cdots, S
 \end{equation}

 Subsequently, the data will be processed in a certain batch size
 and be shuffled to introduce randomness during training.
 This comprehensive preparation ensures the data is well-structured
 and normalized, supporting the training of PH-VAE models effectively.

\subsection{PH-VAE Architecture}

 The proposed PH-VAE framework is
 an extension of the original VAE ~\citep{kingma2014, do2016},
 and the reformulated input data is a hierarchical form of parallel data sequence. By doing so, we can
 process multi-scale polynomial features
 that share a same decoder across these inputs as shown in Figure 1.
 This architecture can be particularly efficient in scenarios where
 the available training data is limited.
 Now we dive into the formulation and architecture in the context of PH-VAE.

 Each exponent \( s \) in dataset \( \boldsymbol{x}_n^s \) (\( s = 1, \dots, S \))
 accords with an encoder. The encoder is composed of a fully connected layer followed
 by an activation function, resulting in the following formula,

 \begin{equation}
 \boldsymbol{h}^s = g\left(\boldsymbol{W}^s \boldsymbol{x}_n^s + \boldsymbol{b}^s \right),
 \end{equation}
where \(\boldsymbol{g}(\cdot)\) is the activation function; \(\boldsymbol{W}^s \), \(\boldsymbol{b}^s \)
are the undetermined weight matrix and bias vector for the $s$-th encoder, respectively; \(\boldsymbol{h}^s \)
represents the output vector of the encoder.

 \bigskip
\begin{figure}[h]
		\begin{center}
			\includegraphics[width=5.5 in]{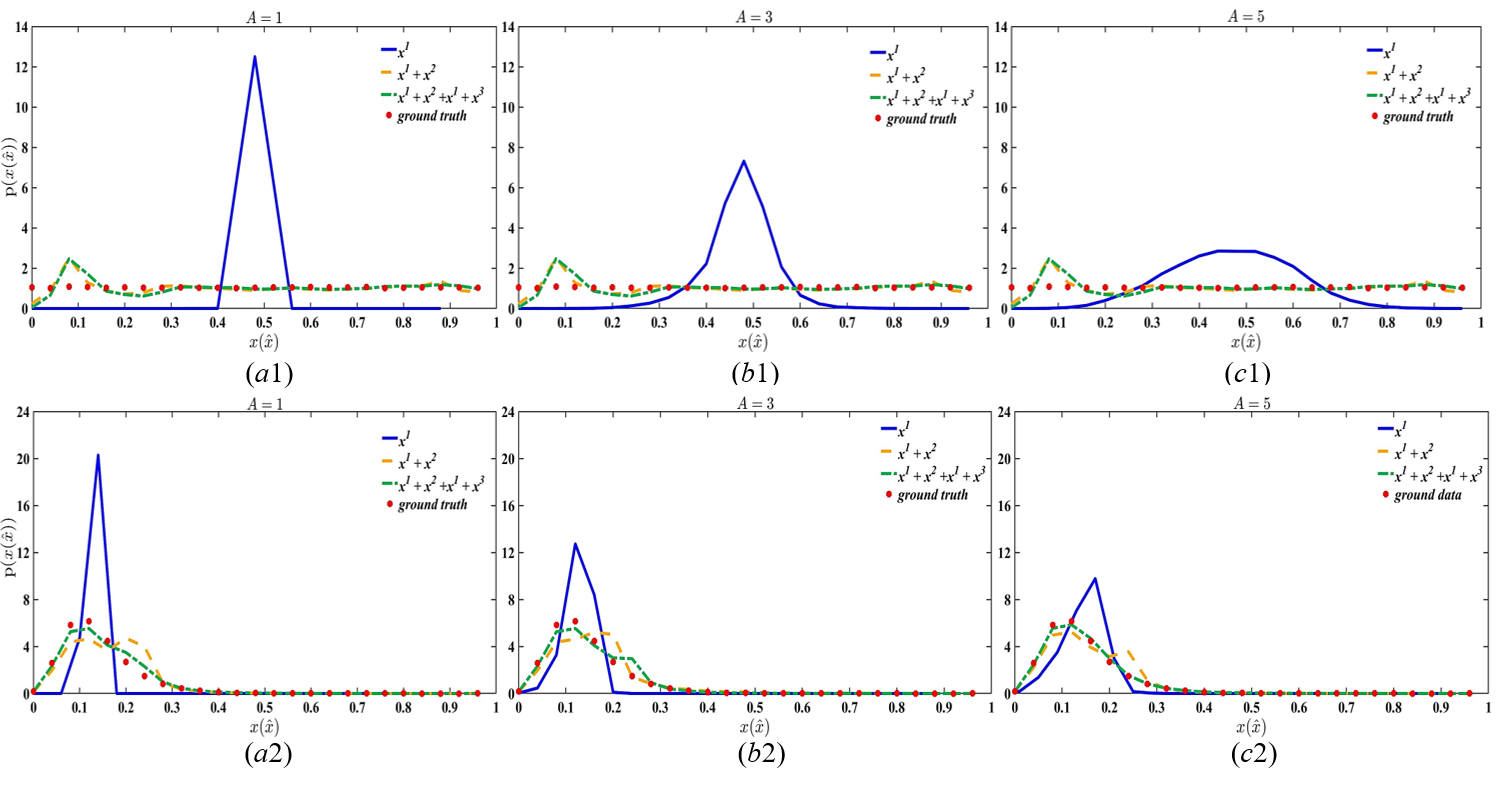}
		\end{center}
	\caption{Probability densities of the predicted results
 for PH-VAE, VAE and the ground truth samples. (a1) Uniform distribution: $A$=1; (b1) Uniform distribution: $A$=3; (c1) Uniform distribution: $A$=5; (a2) Log-normal distribution: $A$=1; (b2) Log-normal distribution: $A$=3; (c2) Log-normal distribution: $A$=5.}
	\label{fig:concrete-fig1a}
\end{figure}

 With the utilization of \( \boldsymbol{h}^s \), two separated fully connected layers are applied
 to compute the mean \( \boldsymbol{\mu}_s \) and the logarithm of the variance \( \log (\boldsymbol{\sigma}_s)^2 \)
 for the latent variable \( \boldsymbol{z} \) as follows,

 \begin{eqnarray}
 \boldsymbol{\mu}_s &=& \boldsymbol{W}_\mu \boldsymbol{h}^s + \boldsymbol{b}_\mu,
 \\
 \log \boldsymbol{\sigma}_{s}^{2} &=& \boldsymbol{W}_\sigma \boldsymbol{h}^s + \boldsymbol{b}_\sigma,
 \end{eqnarray}

\noindent
 where $ \boldsymbol{W}_\mu , \boldsymbol{W}_\sigma $
 are the weight vectors for the respective transformations; \( \boldsymbol{b}_\mu, \boldsymbol{b}_\sigma \) are the bias parameters.
 Then the latent variable \( \boldsymbol{z} \) is sampled from the Gaussian distribution
 through the reparameterization trick, allowing to express $\boldsymbol{z} $ as,

 \begin{equation}
 \boldsymbol{z} = \boldsymbol{\mu} + A \cdot \boldsymbol{\epsilon} \cdot \boldsymbol{\sigma} ,
 \end{equation}
where \( \boldsymbol{\epsilon} \) is a vector sampled from the standard normal
distribution \( \mathcal{N}(0,1) \), introducing noises into the latent variable components,
 \[
 \boldsymbol{\mu} =  {1 \over S} \sum_{s=1}^{S} \boldsymbol{\mu}_s ~~~{\rm and}~~
  \boldsymbol{\sigma}^2 =  {1 \over S} \sum_{s=1}^{S} \boldsymbol{\sigma}_s^2
  \]
   as the amplitude of $\boldsymbol{\epsilon} \cdot \boldsymbol{\sigma}$,
 $A$ is exploited to enhance the neural network's expressiveness and allows for better representation of complex latent space structures beyond independent scaling.

Then the latent variable \( \boldsymbol{z} \) is fed into the decoder layer to reconstruct the original input \( \boldsymbol{x}_n^1 \), enabling the decoder to use a unified representation of the inputs while preserving their shared latent features. It can be defined as,
 \begin{equation}
 \hat{\boldsymbol{x}}_n^1 = f(\boldsymbol{W} \boldsymbol{z} + \boldsymbol{b}),
 \end{equation}
 where \( f(\cdot) \) is the activation function; \( \boldsymbol{W} , \boldsymbol{b} \)
 are the weight matrix and bias vector to be determined in the decoder layer, respectively.

\bigskip
\begin{figure}[h]
		\begin{center}
			\includegraphics[width=5.8 in]{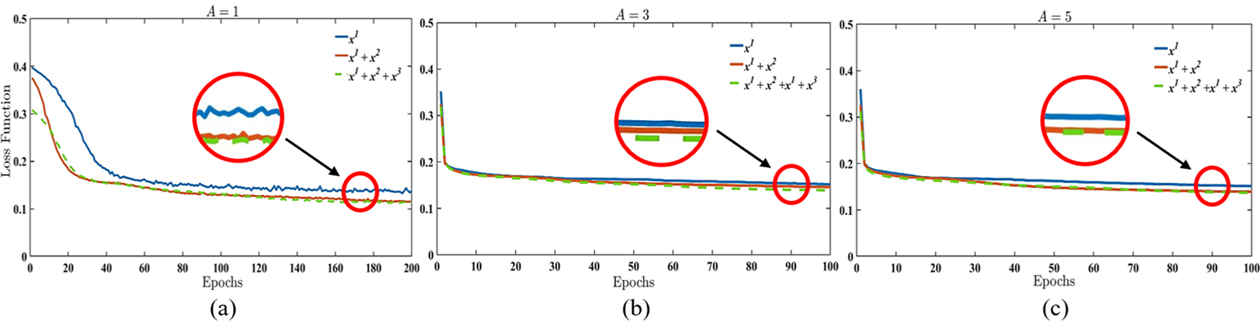}
		\end{center}
	\caption{Values of the loss function in each epoch for the polynomial features $\boldsymbol{x}^s$.
    ($a1$) Uniform distribution: $A$=1; ($b1$) Uniform distribution: $A$=3; ($c1$) Uniform distribution:
     $A$=5; ($a2$) Log-normal distribution: $A$=1; ($b2$) Log-normal
     distribution: $A$=3; ($c2$) Log-normal distribution: $A$=5.}
	\label{fig:concrete-fig1a}
\end{figure}

\subsection{Training and optimization}

 In the proposed PH-VAE, a structured generation process is constructed,
 where the observed data $\boldsymbol{x}_n^1$,
 and the latent variables $\boldsymbol{z}$ are jointly distributed as follows,
 \begin{equation}
 p(\boldsymbol{x}, \boldsymbol{z}) \approx p(\boldsymbol{x}| \boldsymbol{z}) \cdot {1 \over S} \sum_{s=1}^{S}
  p(\boldsymbol{z} | \boldsymbol{x}^{s}),
 \label{eq:IWAE}
 \end{equation}
 where $p(\boldsymbol{x}| \boldsymbol{z})$ indicates the likelihood function
 that models the data reconstruction from $\boldsymbol{z}$, and $p(\boldsymbol{z} | \boldsymbol{x}^{s})$
 denoting the hierarchical prior, capturing the conditional dependency of $\boldsymbol{z}$ on $\boldsymbol{x}^{s}$.

 \bigskip
\begin{figure}
\begin{center}
\begin{minipage}{0.98\linewidth}
		\begin{center}
		   \includegraphics[width=5.5 in]{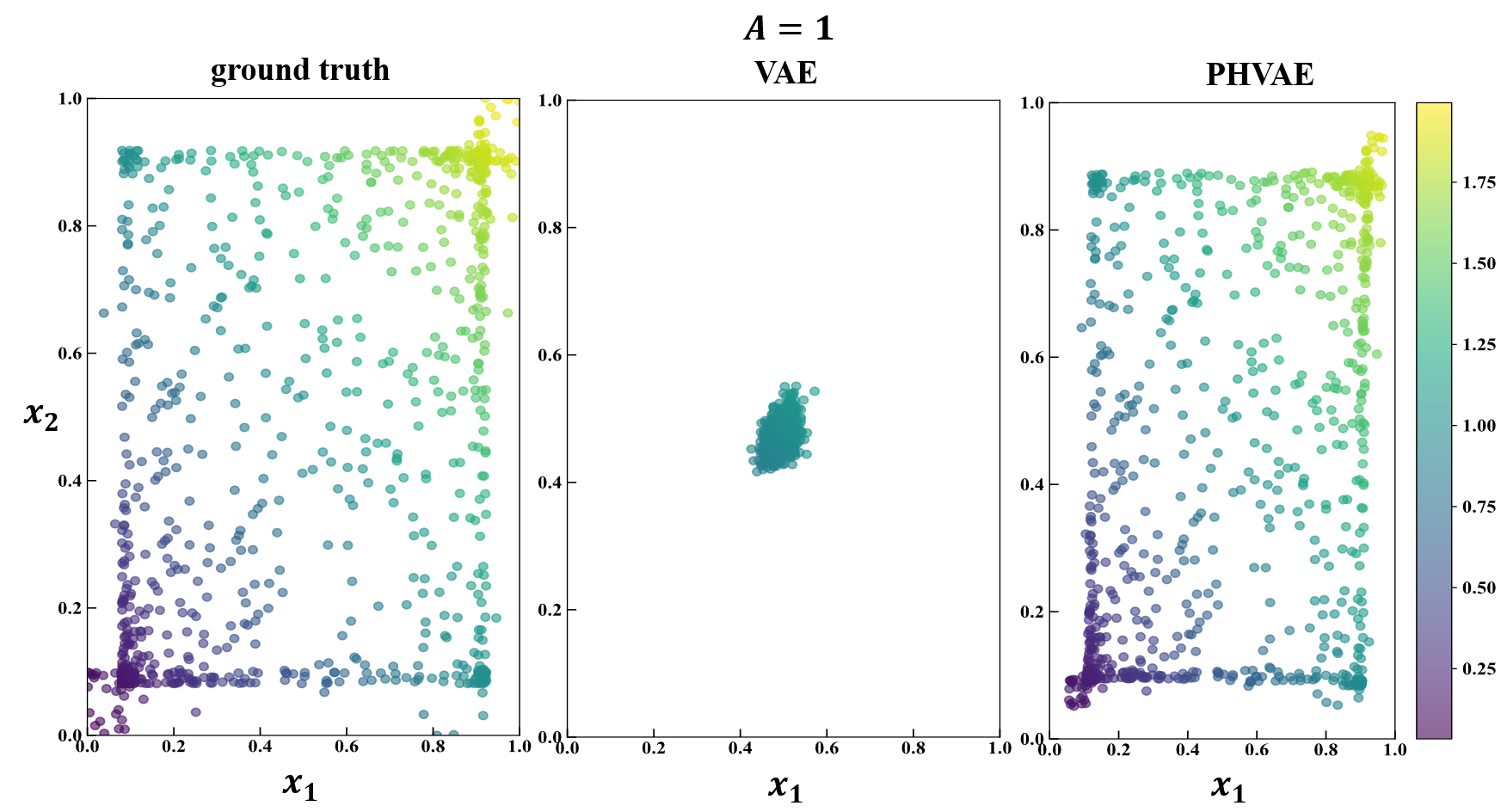}
		\end{center}
	\centering (a)
\end{minipage}
\begin{minipage}{0.48\linewidth}
		\begin{center}
             \includegraphics[width=2.5in]{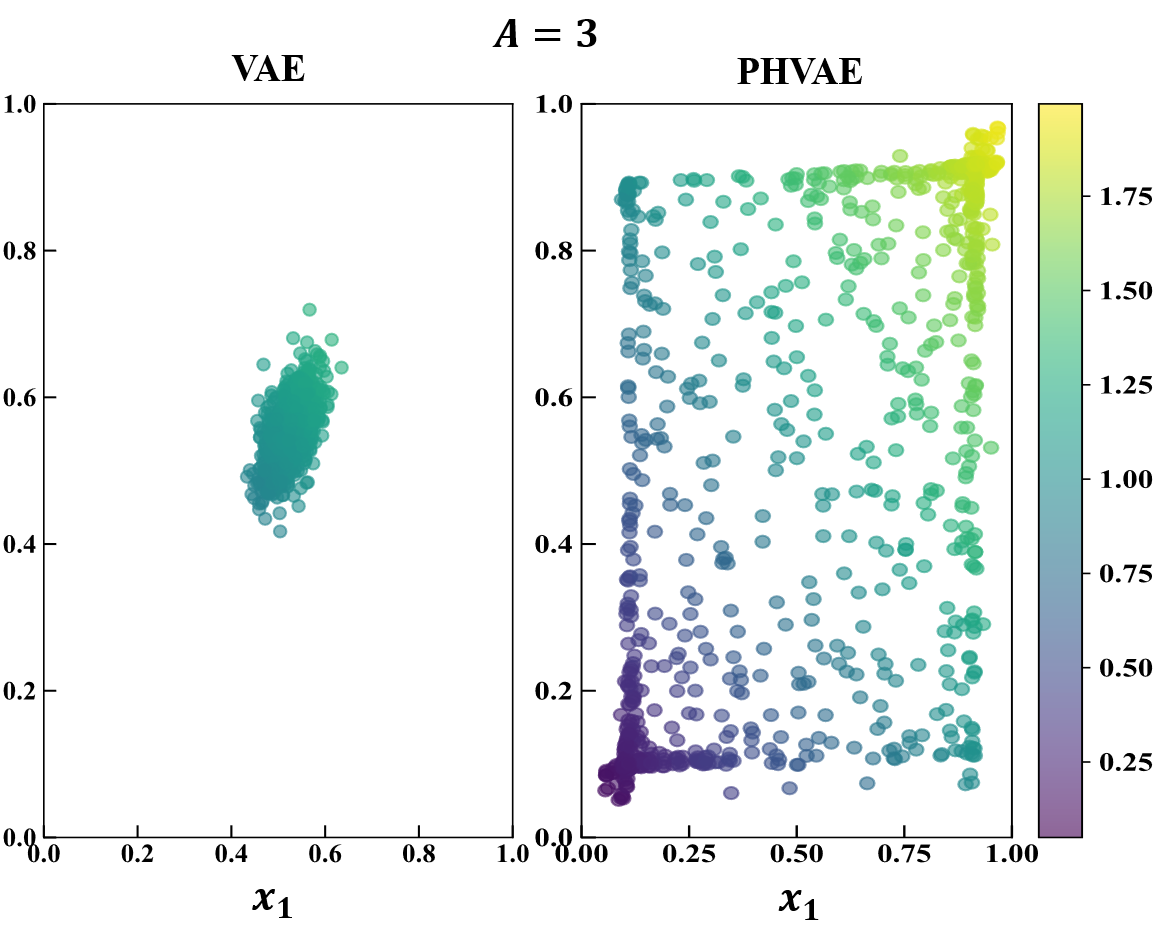}
		\end{center}
	\centering (b)
\end{minipage}
\begin{minipage}{0.48\linewidth}
		\begin{center}
             \includegraphics[width=2.5in]{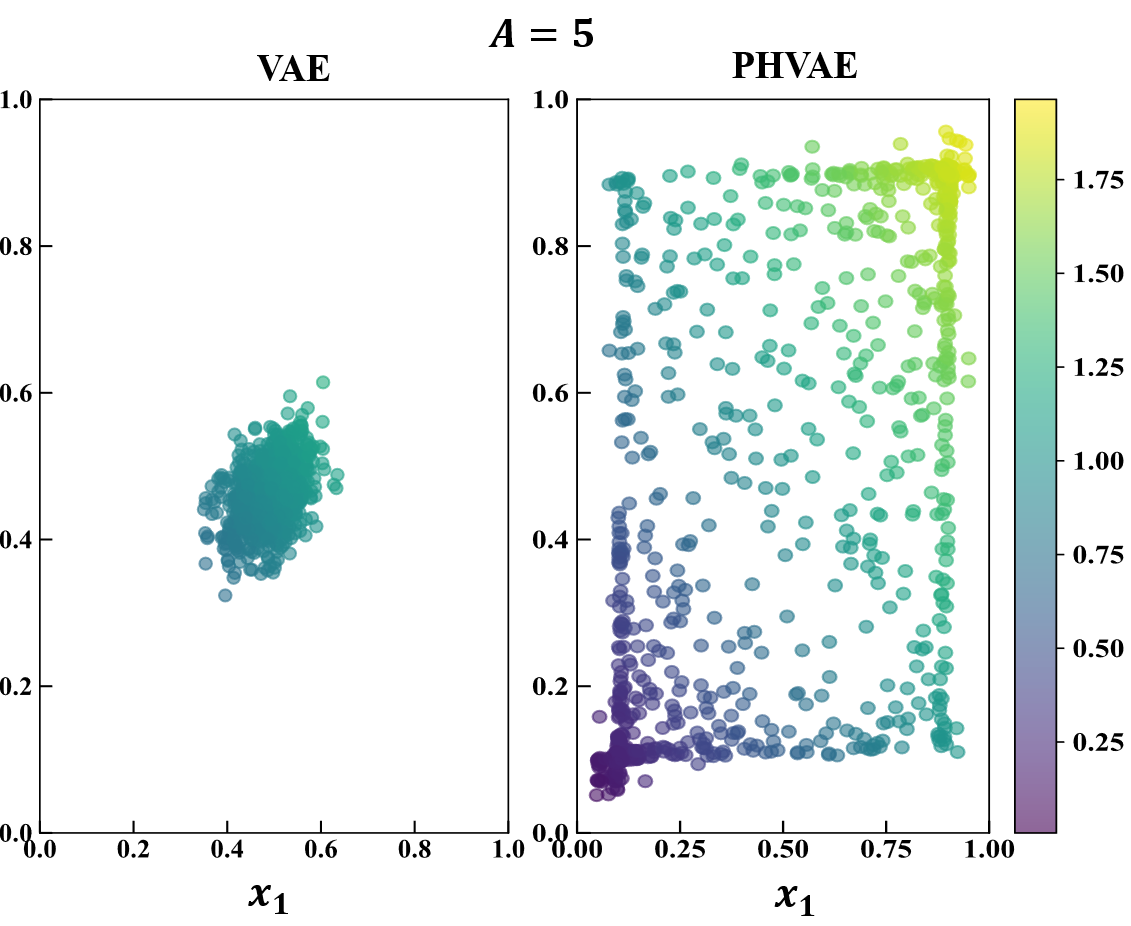}
		\end{center}
	\centering (c)
\end{minipage}
\end{center}
\caption{Comparison of the ground truth data and the reconstructed
results from VAE and PHVAE in the first failure case. (a) $A=1$; (b) $A=3$, and (c) $A=5$.}
\label{fig:concrete-fig1a}
\end{figure}
\medskip

We can derive Eq.~(\ref{eq:IWAE}) from
the Importance Weighted Autoencoder (IWAE) metric. For simplicity, the subscript $n$ of $\boldsymbol{x}_n^1$
has been removed during the derivation.
The exact posterior distribution $p(\boldsymbol{x} | \boldsymbol{z})$ is generally intractable
due to the complexity of the hierarchical dependencies. To solve this problem,
a surrogate model $q(\boldsymbol{z} | \boldsymbol{x})$ is introduced as an approximation,
which is expressed as follows,

\begin{equation}
q(\boldsymbol{z} | \boldsymbol{x}) := {1 \over S} \sum_{s=1}^{S} q_s(\boldsymbol{z} | \boldsymbol{x}^{s}),
\label{eq:Surrogate}
\end{equation}
 where $q_1(\boldsymbol{z} | \boldsymbol{x})$ encodes the dependency of the first latent layer on the input data $\boldsymbol{x}^1$, and $q_s(\boldsymbol{z} | \boldsymbol{x}^{s}), s=2, \cdots S$ capture
 the conditional dependency between successive latent layers.
 This factorization mirrors the hierarchical structure of the generative model,
 facilitating efficient computation of the variational posterior and hence
 a disentangled representation learning.
 Note that $q({\boldsymbol{z}} | {\boldsymbol{x}}) \not = q_1({\boldsymbol{z}}|{\boldsymbol{x}})$.

 In this work, we propose the following evidence lower bound function,
 \begin{equation}
 \mathcal{L}^{PH}_{\text{ELBO}} := \mathbb{E}_{q(\boldsymbol{z} | \boldsymbol{x})} \left[\log p(\boldsymbol{x} | \boldsymbol{z})\right] - {1 \over S}
  \sum_{s=1}^{S} \text{KL}\left[q_s(\boldsymbol{z} | \boldsymbol{x}^{s}) \| p(\boldsymbol{z})\right],
 \label{eq:LoosF}
 \end{equation}
 where, by definition,
 the Kullback-Leibler (KL) Divergence for each layer is given as
 \begin{equation}
 \text{KL}\left[q_s(\boldsymbol{z} | \boldsymbol{x}^{s}) \| p(\boldsymbol{z})\right] =
 \int_{\boldsymbol{z}} \Bigl(
 q_s(\boldsymbol{z} | \boldsymbol{x}^{s})
 \log \Bigr(
 {q_s(\boldsymbol{z} | \boldsymbol{x}^{s}) \over
 p(\boldsymbol{z})}
 \Bigr) d {\boldsymbol{z}}
 \label{eq:KLlayer}.
 \end{equation}

 We define the average of
 the Kullback-Leibler (KL) Divergences of all layers as
 the Polynomial Hierarchical Divergence, i.e.
 \begin{equation}
 {\rm PH} (q({\boldsymbol{z}}|{\boldsymbol{x}})\|p ({\boldsymbol{z}})):=
 {1 \over S} \sum_{s=1}^{S} \text{KL}\left[q_s(\boldsymbol{z} | \boldsymbol{x}^{s}) \| p(\boldsymbol{z})\right]~.
  \label{eq:PL}
 \end{equation}

One can verify the following facts for the measure
${\rm PH} (q({\boldsymbol{z}}|{\boldsymbol{x}})\|p ({\boldsymbol{z}}|{\boldsymbol{x}}))$:
\[
{\rm PH} (q({\boldsymbol{z}}|{\boldsymbol{x}})\|q({\boldsymbol{z}}|{\boldsymbol{x}})) =0~,
\]
if
\[
q_1({\boldsymbol{z}}|{\boldsymbol{x}})=q_2({\boldsymbol{z}}|{\boldsymbol{x}^2})= \cdots =
q_S({\boldsymbol{z}}|{\boldsymbol{x}^S})= q({\boldsymbol{z}}|{\boldsymbol{x}})  =p(\boldsymbol{z}|{\boldsymbol{x}});
\]
or
\[
{1 \over S} \sum_{s=1}^S \mathbb{E}_{q_s(\cdot|{\boldsymbol{x}}^s)}[\log (q_s({\boldsymbol{z}}|{\boldsymbol{x}^s}) )]
= \mathbb{E}_{q(\cdot|{\boldsymbol{x}})} [\log p({\boldsymbol{z}}|{\boldsymbol{x}}) ]~.
\]
Moreover, since
\[
{\rm PH} (q({\boldsymbol{z}}|{\boldsymbol{x}})\|p ({\boldsymbol{z}}))>0, ~\forall q , p >0~,
\]

\noindent
thus, the measure ${\rm PH} (q({\boldsymbol{z}}|{\boldsymbol{x}})\|p ({\boldsymbol{z}})$ may be
viewed as a Divergence, at least a generalized weak divergence.

We note in passing that by the standard derivation procedure one can show that
\begin{eqnarray}
\int_{\boldsymbol{z}} q_s ({\boldsymbol{z}}| {\boldsymbol{x}}^s) \log {
q_s({\boldsymbol{z}}| {\boldsymbol{x}}^s) \over p({\boldsymbol{z}}, {\boldsymbol{x}})
} d{\boldsymbol{z}} &=&
\mathbb{E}_{q_s ({\boldsymbol{z}}| {\boldsymbol{x}}^s)} \Bigl[
\log {
q_s({\boldsymbol{z}}| {\boldsymbol{x}}^s) \over p({\boldsymbol{z}}, {\boldsymbol{x}})
}
\Bigr]
\nonumber
\\
&=&
\mathbb{E}_{q_s ({\boldsymbol{z}}| {\boldsymbol{x}}^s)} \Bigl[
\log q_s({\boldsymbol{z}}|{\boldsymbol{x}}^s) - \bigl(
\log p({\boldsymbol{x}}|{\boldsymbol{z}}) + \log (p({\boldsymbol{z}}))
\bigr)
\Bigr].
\end{eqnarray}

Hence,
\begin{eqnarray}
\mathcal{L}^{PH}_{ELBO} &=&- {1 \over S}\sum_{s=1}^{S}
\left \{
\int_{\boldsymbol{z}} q_s ({\boldsymbol{z}}| {\boldsymbol{x}}^s) \log {
q_s({\boldsymbol{z}}| {\boldsymbol{x}}^s) \over p({\boldsymbol{z}}, {\boldsymbol{x}})
} d{\boldsymbol{z}}
\right \}
\nonumber
\\
&=& {1 \over S} \sum_{s=1}^S \mathbb{E}_{q_s ({\boldsymbol{z}}| {\boldsymbol{x}}^s)} \Bigl[
-\log q_s({\boldsymbol{z}}|{\boldsymbol{x}}^s) + \bigl(
\log p({\boldsymbol{x}}|{\boldsymbol{z}}) + \log (p({\boldsymbol{z}}))
\bigr)
\Bigr]
\nonumber
\\
&=&
\mathbb{E}_{q(\boldsymbol{z} | \boldsymbol{x})}
\left[\log p(\boldsymbol{x} | \boldsymbol{z})\right] - {1 \over S}
\sum_{s=1}^{S} \text{KL}\left[q_s(\boldsymbol{z} | \boldsymbol{x}^{s}) \| p(\boldsymbol{z})\right]~.
\end{eqnarray}
In the last line, we used the definition of Eq. (\ref{eq:Surrogate}), i.e.
\[
 q(\boldsymbol{z} | \boldsymbol{x}) := {1 \over S} \sum_{s=1}^{S} q_s(\boldsymbol{z} | \boldsymbol{x}^{s})~.
 \]

Following the standard procedure, one can also derive an alternative expression of the ELBO objective as follows,
\begin{equation}
\mathcal{L}_{ELBO}^{PH} =  \log p({\boldsymbol{x}}) - {1 \over S}
\sum_{s=1}^S KL[ q_s ({\boldsymbol{z}}| {\boldsymbol{x}}^s) \| p({\boldsymbol{z}} |
{\boldsymbol{x}})),
\end{equation}
where
\begin{equation}
 KL[ q_s ({\boldsymbol{z}}| {\boldsymbol{x}}^s) \| p({\boldsymbol{z}} |
{\boldsymbol{x}}))= \int_{\boldsymbol{z}} q_s ({\boldsymbol{z}}| {\boldsymbol{x}}^s) \log \Bigl( {
q_s({\boldsymbol{z}}| {\boldsymbol{x}}^s) \over
 p({\boldsymbol{z}} |
{\boldsymbol{x}})} \Bigr) d {\boldsymbol{z}}~,
\end{equation}
which is often called the reverse KL divergence.

From Eq. (\ref{eq:LoosF}), one can see that
 $\text{PH}\left[q(\boldsymbol{z} | \boldsymbol{x}) \| p(\boldsymbol{z})\right]$
 penalizes deviations of the variational posterior from the prior.
 By assuming that each surrogate model is a Gaussian distribution,
 the polynomial divergence has a closed form as follows:
 \begin{equation}
 \text{PH}\left[q(\boldsymbol{z} | \boldsymbol{x})
 \| p(\boldsymbol{z})\right] =
 \frac{1}{2S} \sum_{s=1}^S \left(1 + \log \boldsymbol{\sigma}_s^2(\boldsymbol{x}^s) - \boldsymbol{\mu}_s^2(\boldsymbol{x}^s) - \boldsymbol{\sigma}_s^2(\boldsymbol{x}^s) \right)~.
 \label{eq:PGG}
 \end{equation}

\noindent
It should be noted that the product of two Gaussian variables may not be a Gaussian variable, even though
the product of two Gaussian PDFs is still a Gaussian PDF.
Thus, Eq. (\ref{eq:PGG}) is intrinsically an approximation.

  The negative ELBO is minimized to obtain the undetermined parameters
  $\boldsymbol{W}^s, \boldsymbol{b}^s$, $\boldsymbol{W}_\mu, \boldsymbol{b}_\mu$, $\boldsymbol{W}_\sigma,\boldsymbol{b}_\sigma, \boldsymbol{W}$ and $\boldsymbol{b}$, leading to the following loss function,

 \begin{equation}
 \mathcal{L} = -\mathbb{E}_{q(\boldsymbol{z} | \boldsymbol{x})} \left[\log p(\boldsymbol{x} | \boldsymbol{z})\right] + {1 \over S} \sum_{s=1}^{S} \left[\frac{1}{2} \sum \left(1 + \log \boldsymbol{\sigma}_s^2(\boldsymbol{x}^s) - \boldsymbol{\mu}_s^2(\boldsymbol{x}^s) - \boldsymbol{\sigma}_s^2(\boldsymbol{x}^s) \right)\right],
 \end{equation}

To illustrate computer implementation, we show the segment of Python code in the following.
The loss function with $S=1,2, 3$ are defined and implemented as follows,

\begin{verbatim}
class PHVAE(nn.Module):
    def __init__(self):
        super(PHVAE, self).__init__()
        self.latent_dim = latent_dim
        self.encoder = nn.Sequential(......)
        self.mu = nn.Linear(......)
        self.logvar = nn.Linear(......)
        self.decoder = nn.Sequential(......)
    def reparameterize(self, mu, logvar, A=1):
        std = torch.exp(0.5 * logvar)
        eps = torch.randn_like(std)
        return mu + A * eps * std
    def encode(self, x, A=1):
        x = self.encoder(x)
        mu, logvar = self.mu(x), self.logvar(x)
        return self.reparameterize(mu, logvar, A), mu, logvar
    def forward(self, x1, x1**2, x1**3, A=1, S=3):
        z1, mu1, logvar1 = self.encode(x1, A)
        z2, mu2, logvar2 = self.encode(x1**2, A)
        z3, mu3, logvar3 = self.encode(x1**3, A)
        mu = (mu1 + mu2 + mu3) / S
        logvar = torch.logsumexp(torch.stack([logvar1, logvar2, logvar3]), \
                 dim=0) - math.log(S)
        z_combined = reparameterize(mu, logvar, A)
        recon_x1 = self.decoder(z_combined)
        return recon_x1, mu1, logvar1, mu2, logvar2, mu3, logvar3

def loss_function(recon_x1, x1, z1_mu, z1_logvar, z2_mu, z2_logvar, S=3):
    recon_loss1 = nn.functional.mse_loss(recon_x1, x1, reduction='sum')
    kl_loss1 = -0.5 * torch.sum(1 + z1_logvar - z1_mu.pow(2) - z1_logvar.exp())
    kl_loss2 = -0.5 * torch.sum(1 + z2_logvar - z2_mu.pow(2) - z2_logvar.exp())
    kl_loss3 = -0.5 * torch.sum(1 + z3_logvar - z3_mu.pow(2) - z3_logvar.exp())
    return recon_loss1 + (1*kl_loss1 + 1*kl_loss2 + 1*kl_loss3) / S
\end{verbatim}

 The training process with the loss function proceeds iteratively across epochs. Firstly, the forward propagation is exploited to produce the output predictions and compute the associated loss. Afterward, the backward propagation involves computing the gradients of the loss function with respect to the network parameters, accomplished using the chain rule of differentiation. Consequently, the Adam optimization algorithm is employed to minimize the loss function $\mathcal{L}$, and all parameters are updated step by step until $\mathcal{L}$ converges. The architecture of the PH-VAE is illustrated in Figure 1.

\subsection{Disentanglement representation in PH-VAE}

  The proposed PH-VAE modifies the prior structure to a hierarchical polynomial form with the variational posterior. According to (Mutual Information) MI decomposition, ELBO objective in Eq. (14) can be modified as,

\begin{equation}
    \mathcal{L}_{\text{ELBO}}^{PH} =
     \mathbb{E}_{q(\boldsymbol{z}|\boldsymbol{x})} [\log p(\boldsymbol{x}|\boldsymbol{z})] -
     I(\boldsymbol{x}, \boldsymbol{z}) - {1 \over S} KL[q_1(\boldsymbol{z}|\boldsymbol{x}) \| p(\boldsymbol{z})],
\end{equation}
where
\begin{equation}
I(\boldsymbol{x}, \boldsymbol{z}) := { 1 \over S} \sum_{s=2}^S
 KL[q_s(\boldsymbol{z}|\boldsymbol{x}^s) \| p(\boldsymbol{z})],
\end{equation}

\noindent
 where $I(\boldsymbol{x}, \boldsymbol{z})$ is the mutual information (MI).
 It contains the disentangled parts of higher order scale information.
 Compared to the standard VAE, PH-VAE improves MI by leveraging a hierarchical disentangled prior,
 which alleviates the restrictive assumptions on the latent variable $\boldsymbol{z}$. This allows PH-VAE to encode and preserve more information about the observed data $\boldsymbol{x}$, leading to a richer and more expressive latent representation. Additionally, PH-VAE mitigates the constraints imposed by the KL divergence regularization. By distributing the KL penalty across multiple hierarchical layers, the model prevents excessive compression of the latent space, thereby maintaining a more informative posterior distribution. Furthermore, the hierarchical structure of PH-VAE effectively reduces redundancy among latent dimensions, promoting greater disentanglement and ensuring that individual latent variables
 capture distinct generative factors more effectively.

\bigskip
\bigskip
\begin{figure}[h]
		\begin{center}
			\includegraphics[width=6.0 in]{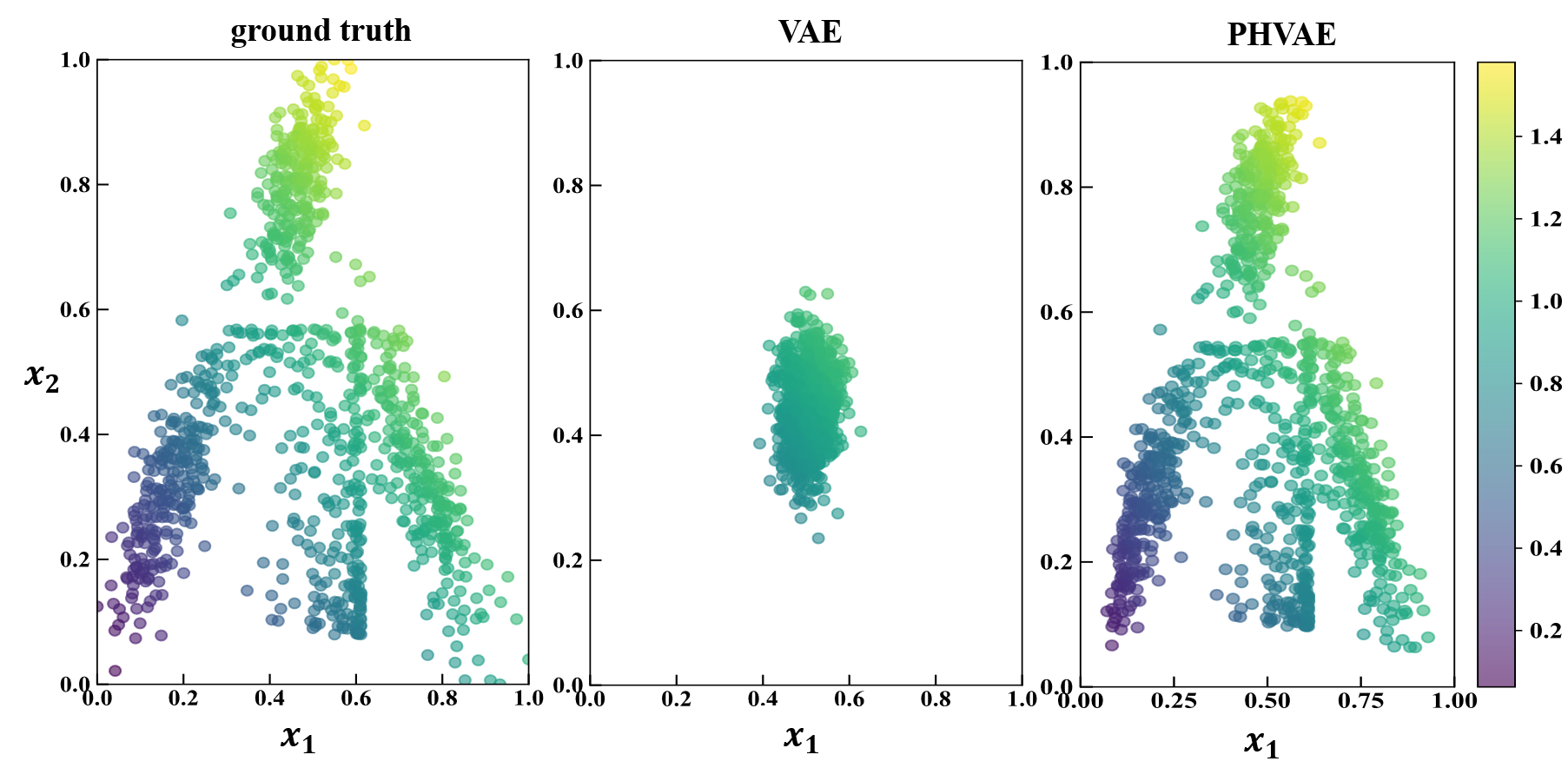}
		\end{center}
	\caption{Comparison of the ground truth data and the corresponding results obtained from VAE
and PH-VAE in the second failure case.}
	\label{fig:concrete-fig1a}
\end{figure}
\bigskip
\bigskip

\section{Experiments}

 To demonstrate the effectiveness and accuracy of PH-VAE, three representative examples are considered, i.e., probability information reconstruction in randomly generated data samples, image information reconstruction in black-and-white digital images, as well as the color images. The outlined procedures are briefly summarized below.

\subsection{Example 1: Probability information reconstruction for data}

 For the training dataset, 20 sets of probabilistic patterns are pre-selected, including log-normal, normal, and uniform distributions, which are utilized to model PH-VAE. It should be noted that each set of distributions contains 50 data points. The raw data is further transformed and normalized as $\{\boldsymbol{x}_n^s\}$ ($n=1,\ldots,20;\ s=1,\ldots,S$). In this case, $S$ is chosen as 3. Thus, PH-VAE possesses three encoders, aiming to introduce diversity and facilitate shared feature learning. $\boldsymbol{x}_n^s$ are then encapsulated as \texttt{TensorDataset} and loaded into \texttt{DataLoader} with a batch size of 5.

 Three encoders map the input data into separate latent space representations, producing both the mean $\boldsymbol{\mu}_s$ and log-variance $\log\boldsymbol{\sigma}_s^2$. The decoder receives a combined latent space representation, computed as the mean from three individual encoders. It is designated to encourage the coupling of information from multiple input streams into a unified latent representation. The dimensions of the input, hidden, and latent layers are specified as 20, 256, and 10, respectively. To be specific, the activation functions $g(\cdot)$ in Eq. (4) and $f(\cdot)$ in Eq. (8) are set to \texttt{ReLU} and \texttt{Sigmoid}, respectively.

\bigskip
\begin{figure}[h]
		\begin{center}
			\includegraphics[width=6.0 in]{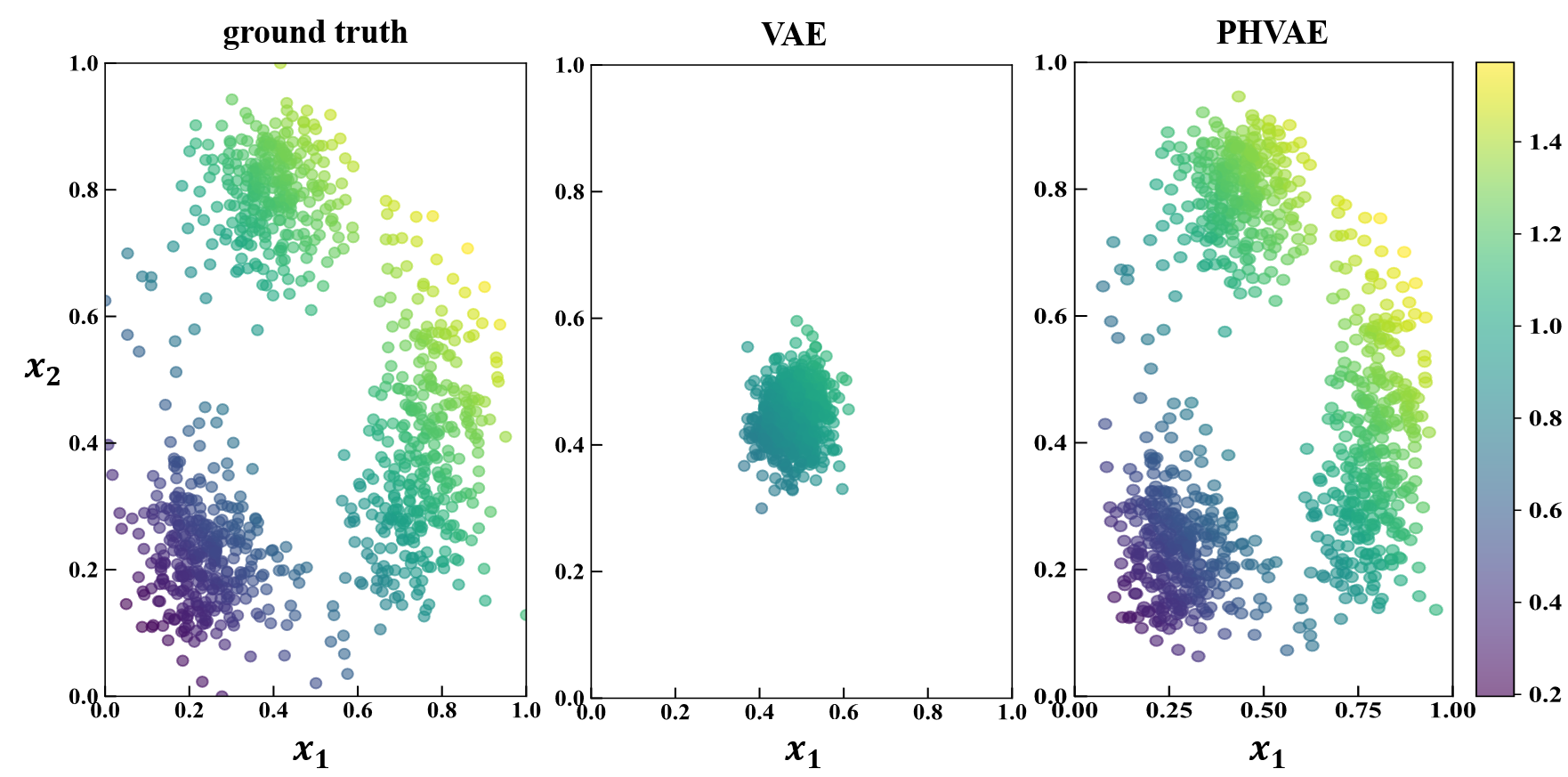}
		\end{center}
	\caption{Comparison of the ground truth data with the corresponding results obtained from VAE and PH-VAE for the cluster example in \citep{yacoby2020}.}
	\label{fig:concrete-fig1a}
\end{figure}

 The loss function in Eq. (14) can be expressed as the sum of the reconstruction loss and polynomial
 divergence losses from all encoders as
 \[
 L = -\mathbb{E}_{q(\boldsymbol{z} | \boldsymbol{x})} [\log p(\boldsymbol{x} | \boldsymbol{z})] +
 {1 \over S}
 \sum_{s=1}^3 \frac{1}{2} \sum \big(1 + \log \boldsymbol{\sigma}_s^2 - \boldsymbol{\mu}_s^2 - \boldsymbol{\sigma}_s^2\big).
 \]

  The Adam optimizer is employed with a learning rate of $5 \times 10^{-4}$. This adaptive method dynamically adjusts the step size for each parameter, improving convergence. During each iteration, the model computes the gradient of the loss with respect to its parameters using back-propagation. The optimizer updates the parameters to minimize the loss. The training process iterates over a predefined number of 100 epochs. The results are provided below.

 Taking log-normal and uniform distributions as the examples, the probability densities of the predicted results from PH-VAE, VAE, as well as the ground truth samples are depicted in Figure 2. The values of $A$ are 1, 3 and 5, which are intended to gradually improve the expressive capacity of VAE. To obtain the smoother curves, the program was executed 100 times to generate a total of 5,000 data points, which were applied to produce Figure 2. It is seen that, for each value of $A$, PH-VAE can better track the hidden probability information compared with VAE in the data, especially for the uniformly distributed data.

 The loss function values with the noise amplifiers $A$=1,3,5 over successive epochs are comprehensively illustrated in Figure 3, providing a detailed visualization of the model's convergence behavior during the training process. It is observed that as the values of $s$ increase, the enriched feature sets introduce additional information into latent space, enhancing its capacity to encapsulate more comprehensive data representations for each $A$. Consequently, the accuracy of the reconstructed probabilistic information improves. When executed on a standard desktop equipped with an Intel\textsuperscript{\textregistered} Core\textsuperscript{TM} i7-10750 CPU @ 2.60 GHz and 16 GB of RAM, the computational time was measured to be 6.9065 seconds.

 According to \citep{yacoby2020}, two specific scenarios under which VAE encounters significant pathologies in accurately learning the true data distribution \(p(\boldsymbol{x})\) resort to posterior intractability and lacking of a simpler likelihood function. These conditions highlight fundamental limitations in the model's ability to balance posterior approximation and likelihood estimation, ultimately impacting its generative performance and density estimation accuracy.

\begin{figure}[h]
		\begin{center}
		\includegraphics[width=6.0 in]{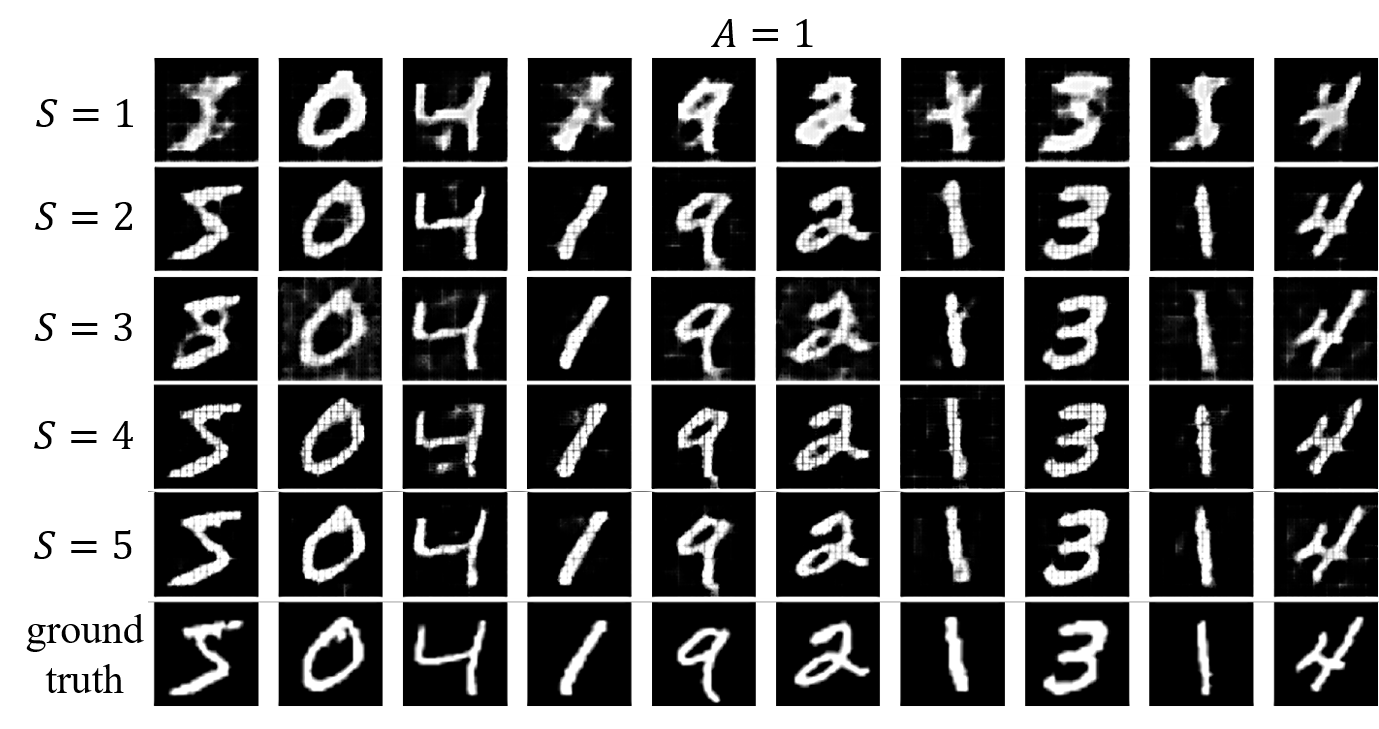}
		\end{center}
	\centering (a)
	\label{fig:concrete-fig1a}
\end{figure}
\bigskip

\begin{figure}[h]
		\begin{center}
		\includegraphics[width=6.0 in]{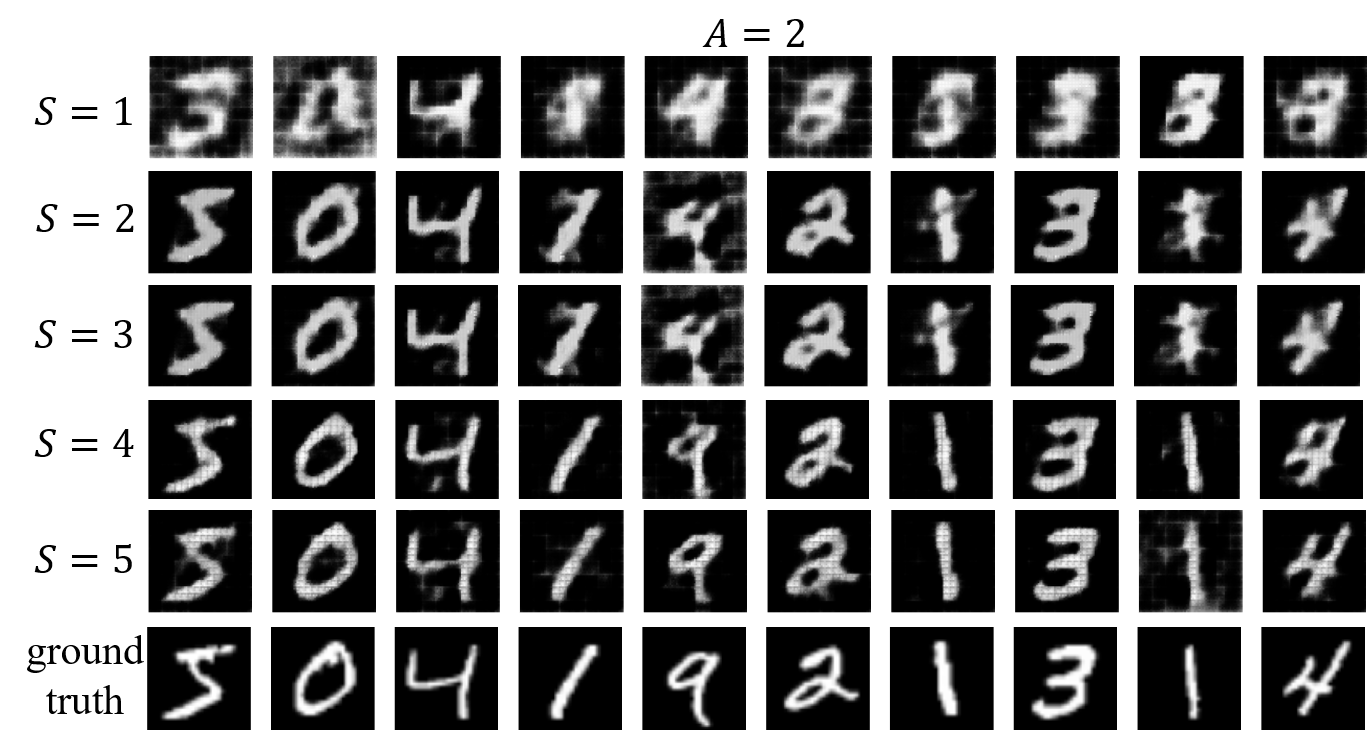}
		\end{center}
        \centering (b)
	\caption{Digital images for VAE, PH-VAE and the ground truth samples. (a) $A$=1; (b) $A=$2.}
	\label{fig:concrete-fig1a}
\end{figure}

 The first condition implies that VAE often relies on simple posterior approximations. However, if the true posterior has complex dependencies, the assumed posterior cannot capture these complexities. This mismatch between the true posterior and the approximating distribution leads to suboptimal learning, as ELBO cannot fully close the gap between the true data likelihood \(p(\boldsymbol{x})\) and the model’s estimate. To verify that PH-VAE is still applicable under this case, Gaussian Mixture Models (GMMs) with nonlinear transformations are adopted to generate the dataset with the mean vectors
 \[
 \begin{array}{l}
 \begin{bmatrix} 3 \\ 3 \end{bmatrix},
 \begin{bmatrix} -3 \\ -3 \end{bmatrix},
 \begin{bmatrix} 3 \\ -3 \end{bmatrix},
 \begin{bmatrix} -3 \\ 3 \end{bmatrix},
 \begin{bmatrix} 0 \\ 0 \end{bmatrix}
 \end{array},
 \]
 The covariance matrices are
 \[
  \begin{array}{cc}
 \begin{bmatrix}
 1 & 0.8 \\
 0.8 & 1
 \end{bmatrix},
 \begin{bmatrix}
 1 & -0.6 \\
 -0.6 & 1
 \end{bmatrix},
 \begin{bmatrix}
 1 & 0.3 \\
 0.3 & 1
 \end{bmatrix},
 \begin{bmatrix}
 0.5 & 0 \\
 0 & 0.5
 \end{bmatrix},
 \begin{bmatrix}
 2 & 1.5 \\
 1.5 & 2
 \end{bmatrix}
 \end{array},
 \]
 where the weights are $ (0.25, 0.25, 0.2, 0.2, 0.1 )$.

 Then the sampled data undergoes hyperbolic tangent \(\tanh(\bm{u})\) and sinusoidal distortion \(0.1 \cdot \sin(2\bm{u})\) for further increasing the complexity of the dataset. The true posterior distribution for such a complex dataset is likely far from the Mean-Field Gaussian assumption commonly used in VAEs. Figure 4(a), (b) and (c) depict the prediction results obtained from VAE and PH-VAE on the aforementioned dataset under varying conditions of $A$, specifically $A$=1,3 and 5. It clearly demonstrates that while the VAE struggles to maintain recognition accuracy under the given conditions, the PH-VAE effectively preserves high recognition accuracy for $A=1,3$ and $5$.

 The second condition indicates that the trade-off between posterior approximation and likelihood modeling results in a compromise that degrades the quality of the learned $p(\boldsymbol{x})$. The Normal distribution basis functions may not be expressive enough to accommodate likelihoods that yield both tractable posteriors and high-quality density estimates. In this case, the dataset includes a mix of correlated, anti-correlated, and narrow-spread Gaussian functions by applying the following mean vectors
 \[
 \begin{bmatrix}
 -4 \\ 0
 \end{bmatrix},
 \quad
 \begin{bmatrix}
 4 \\ 0
 \end{bmatrix},
 \quad
 \begin{bmatrix}
 0 \\ 4
 \end{bmatrix},
 \]
 and the covariance matrices
 \[
 \begin{bmatrix}
 1 & 0.8 \\
 0.8 & 1
 \end{bmatrix},
 \quad
 \begin{bmatrix}
 1 & -0.8 \\
 -0.8 & 1
 \end{bmatrix},
 \quad
 \begin{bmatrix}
 0.5 & 0.3 \\
 0.3 & 0.5
 \end{bmatrix},
 \]
 respectively, as well as a nonlinearly transformed component. The addition of noise and sinusoidal
 distortion \(0.1 \cdot \boldsymbol{\sin(u)}\) mimics real-world complexities,
 making it suitable for testing the robustness
 of generative models like VAEs.

 The prediction results produced by VAE and PH-VAE under the condition of $A$=5 on the specified dataset
 are given in Figure 5. In contrast, the PH-VAE demonstrates superior performance,
 consistently achieving and preserving a high level of recognition accuracy,
 highlighting its robustness and effectiveness in addressing the identified limitations of the VAE.

 Moreover, this paper examines the cluster example presented in reference \citep{yacoby2020}. Most importantly, \citep{yacoby2020} has rigorously deduced the inability of VAE to accurately predict the data in this specific circumstance. Figure 6 illustrates the identified results of both VAE and PH-VAE concerning the dataset for $A$=5, with the caveat that the network setting parameters may differ from those used in \citep{yacoby2020}. The results clearly prove that the proposed PH-VAE is more effective in identifying the cluster points with high accuracy, which emphasizes the superiority of the proposed method in handling the pathology in VAE.

\subsection{Example 2: Pixel Information Reconstruction for black-and-white Images}

 The samples adopted in this experiment originate from the MNIST dataset, ten series of benchmark dataset comprising grayscale images of handwritten digits. Each image is resized to the dimension of $64 \times 64$, and then converted into tensors for processing through the neural network. The selection of $S$ is 5 concerning image reconstruction, so the training dataset is denoted by $\{\boldsymbol{x}_n^s \}$ $(n=1,\cdots,64 \times 64; \; s=1,\cdots,5)$. The data pipeline confirms the uniformity of input data while preserving the original features. The batch size is set to 10, ensuring that all 10 images in the subset are loaded simultaneously for both training and testing.

 The PH-VAE model contains five encoders and a combined decoder. The encoder employs three convolutional layers to upscale the feature dimensions as $32$, $64$, and $128$ channels, each followed by ReLU activations. However, it introduces five distinct latent representations derived from five polynomial features $\boldsymbol{x}, \boldsymbol{x}^2, \boldsymbol{x}^3, \boldsymbol{x}^4, \boldsymbol{x}^5$. The latent vectors are aggregated as
 $\boldsymbol{z}$ before being passed to the decoder.
 The latent space dimensionality is set to $40$.
 A reparameterization trick is used to sample from the latent distribution, to enable
 stochastic differentiation.
 The decoder reconstructs ten original images from the latent representation through a fully connected layer, followed by three transposed convolutional layers. The convolutional layers downscale the feature maps to restore the spatial resolution as $128$, $64$, and $32$. A final Sigmoid activation ensures the output pixel values are constrained within $[0, 1]$.

 The reconstruction loss is calculated for the combined latent representation, while the polynomial divergence terms are computed for each input's latent representation. It can be derived that the expression of the total loss is
 \[
 \mathcal{L} = -\mathbb{E}_{q(\boldsymbol{z}|\boldsymbol{x})} [\log p(\boldsymbol{x}|\boldsymbol{z})]
 + {1 \over 5} \sum_{s=1}^5 \frac{1}{2} \sum \left(1 + \log \boldsymbol{\sigma}_s^2 - \boldsymbol{\mu}_s^2
 - \boldsymbol{\sigma}_s^2 \right).
 \]
 The PH-VAE model is trained over $200$ epochs using the same learning rate. The training process involves computing the loss function and propagating the gradients through both the encoders and decoder via backpropagation. The Adam optimizer is employed to iteratively update the model parameters, including $\boldsymbol{W}^s, \boldsymbol{b}^s, \boldsymbol{W}_\mu, \boldsymbol{b}_\mu, \boldsymbol{W}_\sigma, \boldsymbol{b}_\sigma, \boldsymbol{W}$, and $\boldsymbol{b}$. Throughout each epoch, the loss values are systematically logged to facilitate detailed analysis, monitor convergence, and evaluate the model's performance.

 When $A$=1,2, the superior performance of the proposed PH-VAE architecture is demonstrated by the image results presented in in Figure 7(a) and (b), respectively, with each row corresponding to the polynomial features of $\boldsymbol{x}$ (VAE), $\boldsymbol{x} + \boldsymbol{x}^2$, $\boldsymbol{x} + \boldsymbol{x}^2 + \boldsymbol{x}^3$, $\boldsymbol{x} + \boldsymbol{x}^2 + \boldsymbol{x}^3 + \boldsymbol{x}^4$, $\boldsymbol{x} + \boldsymbol{x}^2 + \boldsymbol{x}^3 + \boldsymbol{x}^4 + \boldsymbol{x}^5$ and the ground truth images, respectively. It can be concluded that augmenting the number of polynomial features can significantly improve the accuracy of image reconstruction for each value of $A$, demonstrating the effectiveness of PH-VAE in capturing complex data patterns.

 In Figure 8, the values of loss function for each polynomial feature are depicted in case of $A$=1,2. As illustrated, when the loss function is stable, the incorporation of additional polynomial features leads to the reduction in the loss function's value, highlighting the effectiveness of augmenting the feature set in improving model performance. For the black-and-white image identification, the total calculation time is measured to be 22.5145 seconds.

\bigskip
\begin{figure}[h]
		\begin{center}
		\includegraphics[width=5.8 in]{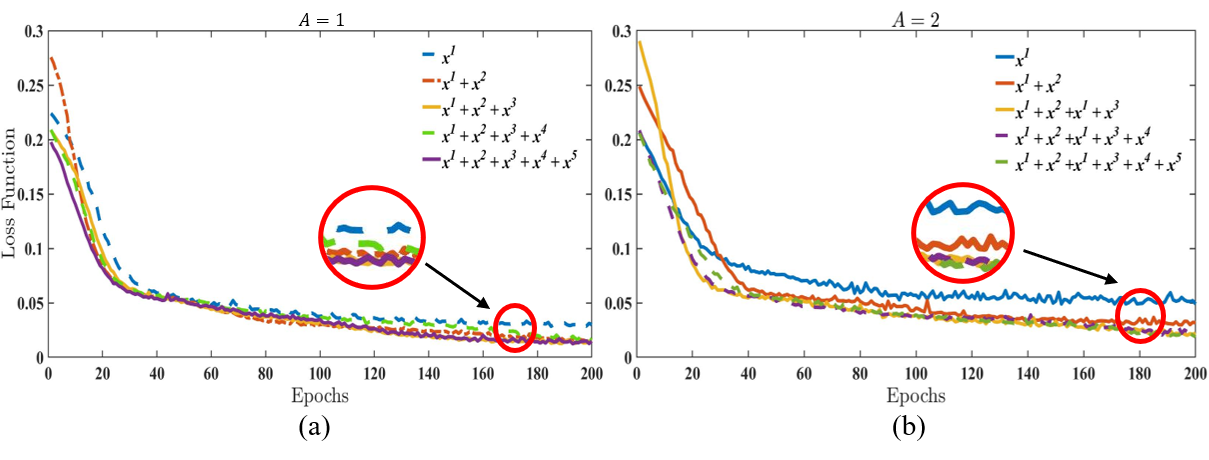}
		\end{center}
	\caption{Values of the loss function in each epoch for the polynomial features $\boldsymbol{x}$ (VAE), $\boldsymbol{x} + \boldsymbol{x}^2$, $\boldsymbol{x} + \boldsymbol{x}^2 + \boldsymbol{x}^3$, $\boldsymbol{x} + \boldsymbol{x}^2 + \boldsymbol{x}^3 + \boldsymbol{x}^4$, and $\boldsymbol{x} + \boldsymbol{x}^2 + \boldsymbol{x}^3 + \boldsymbol{x}^4 + \boldsymbol{x}^5$. (a) $A$=1; (b) $A$=2.}
	\label{fig:concrete-fig1a}
\end{figure}

\bigskip

\subsection{Example 3: Pixel information reconstruction for colorful images}

 Applying a subset of 10 images from the CelebA dataset \citep{Liu2015}, resized to $64 \times 64$ pixels, PH-VAE leverages five polynomial features of the input dataset, namely, $\{\boldsymbol{x}_n^s\}$ $(n=1,\cdots,64 \times 64; s=1,\cdots,5)$, to establish the neural network structure. The \texttt{DataLoader} with a batch size of 10 is used for efficient training. PH-VAE is characterized by five encoders and a decoder. Each encoder consists of four convolutional layers with the dimensions of $64$, $128$, $256$, and $512$, followed by ReLU activation and batch normalization. These layers progressively extract hierarchical features while reducing the spatial dimensions of the input image. The final feature maps are flattened and passed through two fully connected layers to compute the latent mean $\boldsymbol{\mu}_s$ and log-variance $\log \boldsymbol{\sigma}_s^2$.

\bigskip

\begin{figure}
		\begin{center}
			\includegraphics[width=6.0 in]{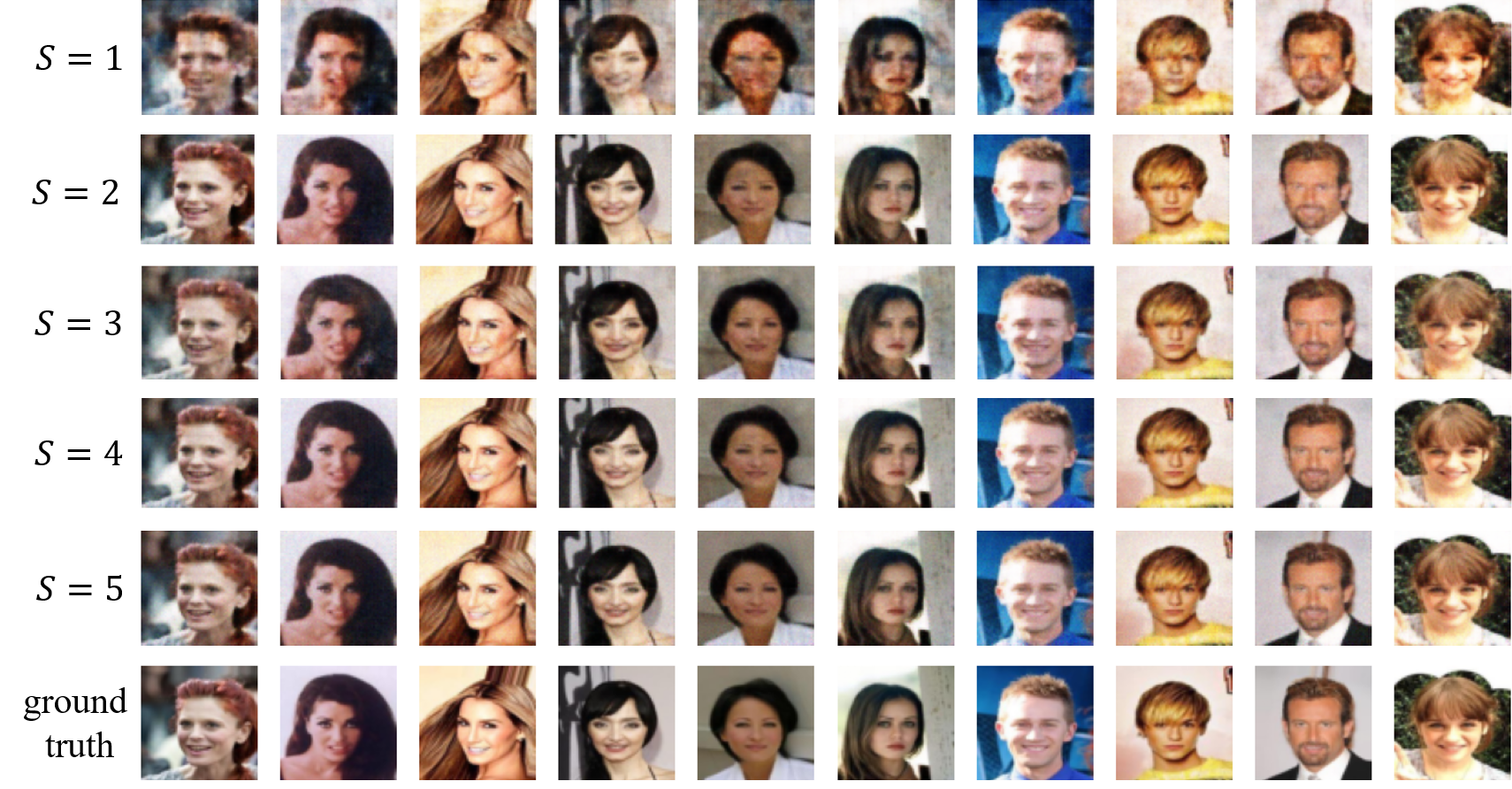}
		\end{center}
	\caption{Facial image comparison among VAE, PH-VAE and the ground truth samples.}
	\label{fig:concrete-fig1a}
\end{figure}

 Through the reparameterization trick, latent vectors $\boldsymbol{z}$ are sampled from a Gaussian distribution defined by these parameters. The decoder reverses this process, starting with a fully connected layer to expand the latent vector into a high-dimensional feature map. It then employs four transposed convolutional layers with dimensions measuring $512$, $256$, $128$, and $64$, each accompanied by ReLU activation and batch normalization in earlier stages, to upsample the feature maps back to the original image dimensions. A sigmoid activation in the final layer ensures the output image has pixel values
 in the range $[0, 1]$.

 The reconstruction loss adheres to the same expression as described in section 3.2. During the training period, the encoders process $\{\boldsymbol{x}_n^s\}$ to generate the respective latent distributions. These latent vectors are combined and passed through the shared decoder to reconstruct the input. The optimizer, typically Adam, minimizes the total loss by backpropagating gradients through both the encoder and decoder. Over $200$ epochs, the model learns to encode robust, integrated features from the input variations, improving the fidelity of the reconstructions and the generalization of the latent space.

 Figure 9 depicts the image reconstruction results to evidence the effectiveness of the proposed PH-VAE architecture for $A$=1. The first row describes the results of VAE; the second row to the fifth one represent the prediction results of polynomial features $\boldsymbol{x} + \boldsymbol{x}^2$, $\boldsymbol{x} + \boldsymbol{x}^2 + \boldsymbol{x}^3$, $\boldsymbol{x} + \boldsymbol{x}^2 + \boldsymbol{x}^3 + \boldsymbol{x}^4$, and $\boldsymbol{x} + \boldsymbol{x}^2 + \boldsymbol{x}^3 + \boldsymbol{x}^4 + \boldsymbol{x}^5$, respectively, and the last row comprises the ground truth samples. These results highlight that increasing the number of polynomial features can enhance the accuracy of image reconstruction to some extent, thereby underscoring the effectiveness of PH-VAE in capturing intricate data patterns.
 The total processing time for colorful image identification is recorded as 85.3892 seconds.

\section*{Conclusions}

The variational autoencoder is a generative AI model in machine learning
that can generate new data in the variant form that is derived
from the input data distributions which the VAE was trained to approximate.
The trained VAE latent space offers exciting opportunities
to develop data-driven out-of-the-box design processes in a creative fashion.
In particular, it may automatically generate multiple novel designs that
are visually evocative, aesthetically related, and reminiscent in appearance
of the input data but that did not exist in training design subspace.

However, the lack of latent space structure of the conventional VAE
prevents us from deliberately controlling the data attributes, e.g. image features,
of newly generated data from the latent space.

To resolve this issue,
in this work, we propose and develop a novel PH-VAE model
 that uses hierarchical input data format (without increase the data size) to
 first disentangle the input information in some degree
 dedicated to reconstructing information in random and image datasets.
 The proposed approach puts forward a novel polynomial divergence that is
 a multiscale generalization
 of the conventional KL divergence and utilizes it in its
 innovative loss function. The purpose of the polynomial divergence provides
 a means to control and to measure disentangled information.
 By leveraging the concept of hierarchical polynomial divergence,
 we are able to integrate the reconstruction loss as well as
 the generation ability
 with the different disentangled parts of the input information as well
 as the disentangled parts of re-construction or generation data,
 which provides a concrete and effective means to disentangle
 the information re-construction and generation, as oppose to
 some implicit disentangled representation methods.

 By addressing the inherent limitations of conventional VAE,
 PH-VAE achieves superior reconstruction accuracy and generative ability,
 while maintaining high computational efficiency.
 Moreover, since the method incorporates higher-order polynomial features
 into the encoder parts, it effectively alleviates the posterior failure phenomenon.
\\

\bigskip

\section*{Statements and Declarations}
The authors declare that there is no financial or non-financial interests that are directly or indirectly related to this
work.

\end{document}